\pdfoutput=1
\documentclass[conference]{IEEEtran}
\IEEEoverridecommandlockouts

\usepackage{cite}
\usepackage{amsmath,amssymb,amsfonts}
\usepackage{algorithmic}
\usepackage{graphicx}
\usepackage{textcomp}
\usepackage{xcolor}
\usepackage{hyperref}
\usepackage{amsthm}
\newtheorem{definition}{Definition}
\usepackage[ruled,linesnumbered]{algorithm2e}
\def\BibTeX{{\rm B\kern-.05em{\sc i\kern-.025em b}\kern-.08em
    T\kern-.1667em\lower.7ex\hbox{E}\kern-.125emX}}

\usepackage{xcolor}
\usepackage{subcaption}

\usepackage{tabularx,booktabs}
\newcolumntype{Y}{>{\centering\arraybackslash}X}

\begin{document}

\newif\ifanonymous

\makeatletter
\newcommand{\newlineauthors}{%
  \end{@IEEEauthorhalign}\hfill\mbox{}\par
  \mbox{}\hfill\begin{@IEEEauthorhalign}
}
\makeatother

\title{Implet: A Post-hoc Subsequence Explainer for Time Series Models}

\ifanonymous

\author{\IEEEauthorblockN{Anonymous}
\IEEEauthorblockA{} \\ \\ \\ }

\else

\author{\IEEEauthorblockN{Fanyu Meng\IEEEauthorrefmark{2}}
\IEEEcompsocitemizethanks{\IEEEcompsocthanksitem\IEEEauthorrefmark{2}equal contribution}
\IEEEauthorblockA{\textit{Department of Computer Science} \\
\textit{UC Davis}\\
Davis, USA \\
fymeng@ucdavis.edu}
\and
\IEEEauthorblockN{Ziwen Kan\IEEEauthorrefmark{2}}
\IEEEauthorblockA{\textit{Department of Computer Science} \\
\textit{UC Davis}\\
Davis, USA \\
zwkan@ucdavis.edu}
\and
\IEEEauthorblockN{Shahbaz Rezaei}
\IEEEauthorblockA{\textit{Department of Computer Science} \\
\textit{UC Davis}\\
Davis, USA \\
srezaei@ucdavis.edu}
\newlineauthors
\IEEEauthorblockN{Zhaodan Kong}
\IEEEauthorblockA{\textit{Department of Computer Science} \\
\textit{UC Davis}\\
Davis, USA \\
zdkong@ucdavis.edu}
\and
\IEEEauthorblockN{Xin Chen}
\IEEEauthorblockA{\textit{College of Engineering} \\
\textit{Georgia Institute of Technology}\\
Atlanta, USA \\
xinchen@gatech.edu}
\and
\IEEEauthorblockN{Xin Liu}
\IEEEauthorblockA{\textit{Department of Computer Science} \\
\textit{UC Davis}\\
Davis, USA \\
xinliu@ucdavis.edu}
}

\fi

\maketitle

\begin{abstract}
Explainability in time series models is crucial for fostering trust, facilitating debugging, and ensuring interpretability in real-world applications. In this work, we introduce \textit{Implet}, a novel post-hoc explainer that generates accurate and concise subsequence-level explanations for time series models. Our approach identifies critical temporal segments that significantly contribute to the model's predictions, providing enhanced interpretability beyond traditional feature-attribution methods. Based on it, we propose a cohort-based (group-level) explanation framework designed to further improve the conciseness and interpretability of our explanations. We evaluate \textit{Implet} on several standard time-series classification benchmarks, demonstrating its effectiveness in improving interpretability. The code is available at \ifanonymous\url{https://github.com/anonymous}\else\url{https://github.com/LbzSteven/implet}\fi.
\end{abstract}


\section{Introduction}

\begin{figure}[t]
\centering
\includegraphics[width=0.75\linewidth]{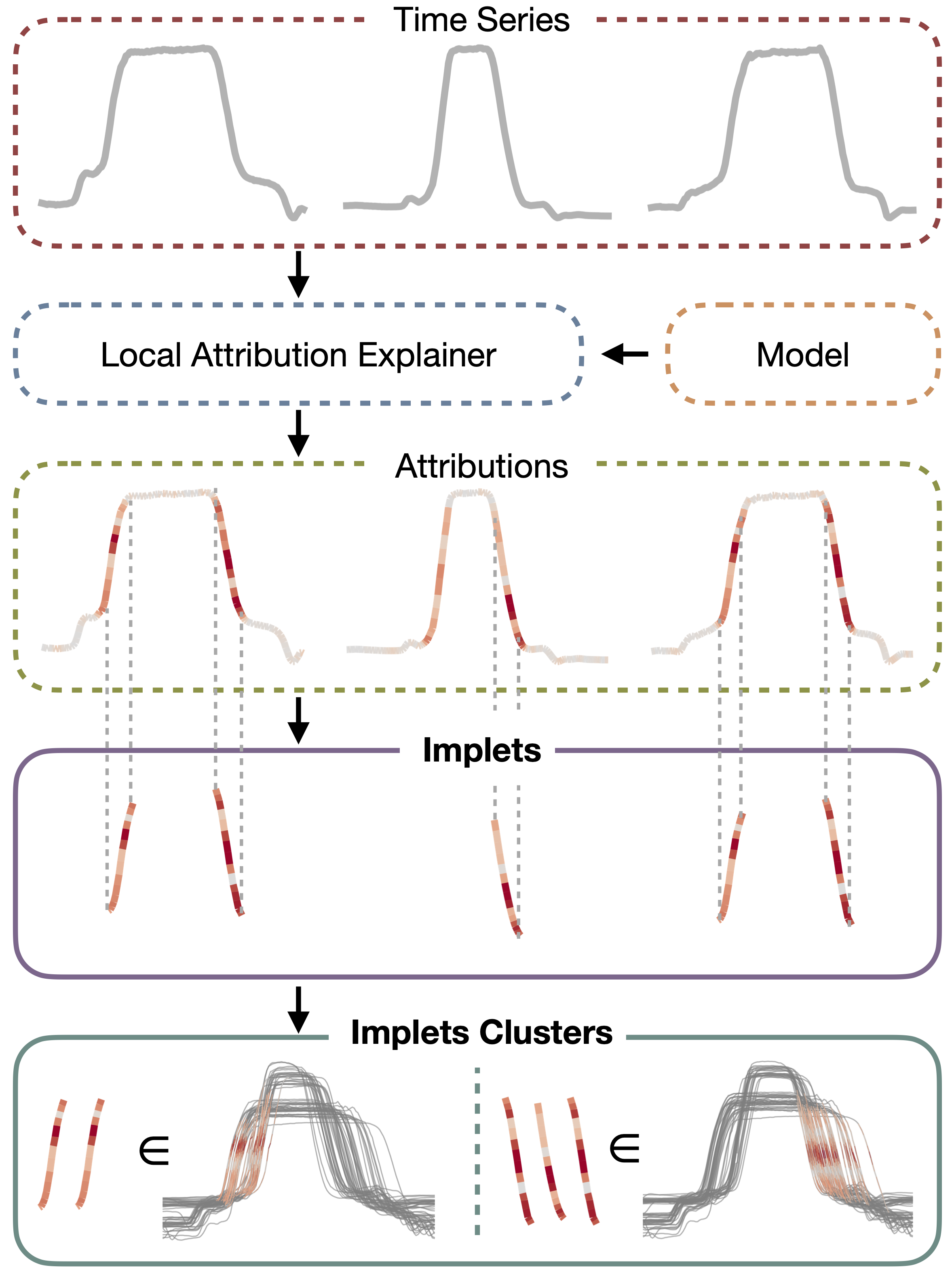}
\caption{The workflow of the proposed post-hoc subsequence explainer \textbf{Implet}. It first computes attributions, then identifies continuous subsequences with consistently high attributions. Lastly, subsequences with similar shapes and attributions are clustered together to enhance conciseness.}
\label{fig:title}
\end{figure}

Deep learning models have demonstrated remarkable success in various time series forecasting and classification tasks, often surpassing traditional statistical methods. Despite their effectiveness, these models are frequently considered \textit{black-boxes}, making their predictions challenging to interpret. Understanding which temporal patterns or subsequences contribute significantly to a model's decisions is crucial for building trust, debugging erroneous predictions, and making informed decisions, especially in high-stakes domains such as finance, healthcare, and climate science.

Existing explainability methods for time series predominantly rely on feature attribution techniques, such as gradient-based saliency maps or perturbation-based approaches. While these methods provide valuable insights into individual time points or features influencing model predictions, their high dimensionality can complicate interpretation.

In comparison, subsequence-based explanations, such as shapelet-based methods, are a type of global explanation that offer more intuitive insights by identifying discriminative temporal patterns within time series data. However, these methods typically inherent instead of post-hoc, and may not match the predictive performance of state-of-the-art architectures like InceptionTime \cite{inceptiontime} and may fail to generalize to complex deep learning models. To bridge this gap, we introduce \textit{Implet} (importance-based shapelets), a post-hoc subsequence explainer that highlights critical temporal segments driving model predictions. Our method effectively combines the interpretability of subsequence-level insights with the flexibility and accuracy of modern deep learning architectures.

To further enhance the interpretability and conciseness of Implet, we introduce a cohort explanation framework. XAI methods are usually categorized into global versus local method. Globals methods are concise, but may not generalize well to all samples; local methods explain each sample, but can be overly verbose. Our cohort explanations strike a balance between the two by grouping similar subsequences into representative clusters, significantly reducing redundancy. This framework enables users to interpret model decisions at a higher abstraction level, making explanations more coherent and actionable for domain experts.

The key contributions of our work are:
\begin{itemize}
\item Introducing \textit{Implet}, a novel post-hoc explainer that identifies salient time series subsequences. It combines the simplicity of subsequence-based explanations and the post-hoc nature of feature attributions. 
\item Proposing \textit{Coh-Implet}, a cohort explanation framework to cluster similar subsequences, thereby improving explanation conciseness. To the best of our knowledge, this is the \textbf{first work} on cohort explanation in time series. Fig.~\ref{fig:title} shows the outline of the proposed framework of implet and its cohort explanation.
\item Designing a novel baseline method for subsequence ablation analysis in time series.
\item Validating Implet across real-world time series datasets, demonstrating its effectiveness in capturing essential predictive factors utilized by deep learning models.
\end{itemize}

\section{Related Work}

\subsection{Time Series Explainability}
\textit{Feature Attribution:} 
Feature attribution methods are post-hoc explainability techniques that quantify the contribution of individual input features or specific time steps toward a model’s predictions. Common gradient-based attribution methods include Grad-CAM \cite{gradcam}, Integrated Gradients \cite{intgrad}, and Input$\times$Gradient \cite{inputxgrad}. Another popular approach is occlusion, which involves masking parts of the input and evaluating changes in the model’s predictions \cite{occlusion}. Similarly, Dynamask \cite{crabbe2021explaining} utilizes dynamic masks to evaluate the importance of timesteps. LIME approximates feature importance through local linear surrogate models \cite{lime}, while SHAP computes feature attributions using cooperative game theory principles, accounting for interactions among features \cite{shap}. These methods are generally model-agnostic and can be readily adapted for time series applications.

More recently, attention mechanisms have also been explored for explainability \cite{attention}, with specific adaptations to time series domains \cite{ts-attention-1, ts-attention-2}. However, such methods are primarily limited to transformer-based architectures.

Additionally, several attribution techniques have been designed for time series data. For instance, \cite{ts-perturb-1} incorporates frequency-domain information, and \cite{ts-perturb-2} employs generative models to produce in-distribution perturbations for attribution analysis.

Our approach leverages feature attribution techniques to identify salient subsequences within time series, simplifying the explanations and enhancing transparency and potentially trust in the underlying model.

\textit{Instance-based Explanations:}
Instance-based explanation methods, such as counterfactual explanations \cite{counterfactual} and prototype selection \cite{prototype}, provide insights by identifying representative or minimally altered instances influencing model predictions. In time series, counterfactual explanations aim to generate minimally perturbed sequences that alter the model's predictions, thus elucidating decision boundaries \cite{NG, COMTE}. However, these methods are often computationally expensive and can be less inherently interpretable to humans. In contrast, our method differs by extracting concise subsequences directly from the original instances, highlighting only essential temporal segments to improve interpretability.

\textit{Subsequence-based Explanations:}
Subsequence-based explanations have gained attention due to their capacity to offer intuitive and interpretable insights by highlighting discriminative temporal patterns within time series data. They usually find discriminative subsequences by maximizing the information gain between samples with and without the subsequence. Prominent approaches, such as shapelet methods, have been widely studied for interpretability in classification and forecasting tasks \cite{shapelet}. Subsequent advancements include incorporating feature selection techniques \cite{shapelet-transform}, multi-channel classification capabilities \cite{ghalwash2012early}, Dynamic Time Warping (DTW) for improved pattern alignment \cite{shapelet-dtw}, and deep learning-based attention mechanisms for enhanced fidelity \cite{shapelet-learning}.

However, most existing shapelet-based methods focus on \textit{inherent} explanations—explanations generated during the model training process—rather than post-hoc explanations. This limits their flexibility, and the models utilizing shapelets often fail to match state-of-the-art deep learning performance. One could train a shaplets model based on the inputs and outputs of a black-box model, but it raises concerns regarding whether the shapelets truly explain the model behavior or merely capture data characteristics. To the best of our knowledge, the only prior work on post-hoc subsequence explainers is LASTS \cite{lasts}, which generates exemplars and counter-exemplars, fits a decision tree, and then uses shapelets to explain the decision branches. In contrast to LASTS, our method directly derives subsequences from local feature attributions, ensuring that explanations faithfully represent the model's internal decision-making rather than the dataset alone.

\begin{algorithm}[t]
\small
\caption{Implet Extraction}
\label{alg:implet_extraction}
\SetKwInOut{Input}{Input}
\SetKwInOut{Output}{Output}
\SetKwComment{Comment}{/* }{ */}
\let\oldnl\nl
\newcommand{\nonl}{\renewcommand{\nl}{\let\nl\oldnl}}
\Input{Sample $\mathbf{x}$, \\ attribution scores w.r.t. class $C$: $\mathbf{w}^C$.}
\Output{Implets w.r.t. class $C$: $\{\mathbf{I}^C_1, \dots, \mathbf{I}^C_r\}$.}
$\mathrm{implets} \gets []$ \\
$i \gets 1$ \\
\Repeat{$i > \mathrm{len}(\mathbf{x}) - \ell_{\min} + 1$}{
  \uIf{$w_i \ge \phi$}{
    $j_{\min} \gets i + \ell_{\min} - 1$ \\
    $j_{\max} \gets \min(i + \ell_{\max} - 1, \mathrm{len}(\mathbf{x}))$ \\
    \nonl \texttt{// find the best end loc} \\
    $j^* \gets \arg\max_{j=j_{\min}}^{j_{\max}} s(i, j; \mathbf{x}, \mathbf{w}^C)$ \\
    \uIf{$s(i, j^*; \mathbf{x}, \mathbf{w}) \ge \phi$}{
      append $\mathbf{I}(i, j^*; \mathbf{x}, \mathbf{w}^C)$ to $\mathrm{implets}$ \\
      \nonl \texttt{// skip to the end of the implet} \\
      $i \gets j^* + 1$ 
    }
    \Else{
      $i \gets i + 1$
    }
  }
  \Else{
    $i \gets i + 1$
  }
}
\textbf{return} $\mathrm{implets}$
\end{algorithm}

\subsection{Cohort Explanation}

Cohort explanations interpret model predictions by analyzing groups (cohorts) of similar instances instead of individual samples. Cohort-based methods reduce variance in explanations, improve conciseness, and capture broader patterns in the model's decision-making process. Typically, cohort-based explanations involve applying local explanation methods to instances and subsequently clustering those instances based on similarities in their explanations.

Notable cohort explanation methods include Glocal-SHAP \cite{glocal-shap}, which computes averaged SHAP values within predefined cohorts. Other methods automatically infer cohort structures; for example, VINE clusters Individual Conditional Expectation (ICE) curves through unsupervised clustering techniques \cite{vine}, while CohEx employs iterative supervised clustering to enhance explanation locality \cite{cohex}. Additionally, REPID \cite{repid} and \cite{molnar2023model} utilize tree-based partitioning methods to form cohorts based on ICE curves or permutation feature importances. The GADGET framework extends these methods by integrating functional decomposition for improved cohort partitioning \cite{gadget}. Beyond cohort partitioning, methods such as GALE refine local explanations (e.g., LIME scores) using homogeneity-based reweighting prior to aggregation, especially useful in multiclass classification settings \cite{gale}.

Despite that these methods can be applied to time series data, their explanations are not ideal: time series samples can have arbitrary lengths, samples may not be aligned, and discriminative information can appear at different locations. Therefore, it's challenging to directly apply the naive sample aggregation method used in the existing cohort explanations. To the best of our knowledge, no direct work has specifically addressed cohort explanations for time series. Our approach explicitly clusters salient subsequences into cohorts, generating group-level explanations that significantly improve clarity, interpretability, and conciseness.

\section{Method}

\begin{algorithm}[t]
\small
\caption{Coh-Implet (Implet Clustering)}
\label{alg:implet_cluster}
\SetKwInOut{Input}{Input}
\SetKwInOut{Output}{Output}
\SetKwComment{Comment}{/* }{ */}
\let\oldnl\nl
\newcommand{\nonl}{\renewcommand{\nl}{\let\nl\oldnl}}
\Input{Implets $\{\mathbf{I}_1, \dots, \mathbf{I}_r\}$.}
\Output{Optimal number of clusters $k^*$, \\ 
        Coh-implets (centroids) $\{\mathbf{M}_1, \dots, \mathbf{M}_{k^*}\}$, \\ 
        Cluster assignments $\{a_1, \dots, a_r\}$.}
$k^* \gets 1$ \Comment*[r]{optimal number of clusters}
$h^* \gets -\infty$ \Comment*[r]{best silhouette score}
$M^* \gets \{\}$ \Comment*[r]{best centroids}
$A^* \gets \{\}$ \Comment*[r]{best assignments}

\For{$k \gets 1, \dots, k_{\max}$}{
  \For{$q \gets 1, \dots, Q$ \Comment*[r]{repeat Q times}}{
    rand. init. centroids $\{\mathbf{M}_1, \dots, \mathbf{M}_k\}$ from implets \\
    \Repeat{convergence}{
      \For{$j \gets 1, \dots, r$}{
        $a_j \gets \arg\min_{c=1}^k d_\mathrm{dtw}(\mathbf{I}_j, \mathbf{M}_c)$
      }
      \For{$c \gets 1, \dots, k$}{
        \nonl \texttt{// DTW Barycenter Averaging}
        $\mathbf{M}_c \gets \mathrm{DBA}(\{\mathbf{I}_j \mid a_j = c\})$
      }
    }
    $h \gets \mathrm{Silhouette}_{\mathrm{dtw}}(\{\mathbf{I}_j\}, \{a_j\})$ \\
    \If{$h > h^*$}{
      update $k^* \gets k$, $h^* \gets h$, $M^* \gets \{\mathbf{M}_c\}$, $A^* \gets \{a_j\}$
    }
  }
}
\textbf{return} $k^*, M^*, A^*$
\end{algorithm}


\subsection{Implet}

The core assumption underlying Implet is that feature attribution methods accurately reflect the behavior of time series models. By ``accurate," we mean that time steps with higher absolute attribution values significantly influence the model’s predictions compared to steps with near-zero attributions. However, raw attributions are often high-dimensional and ``spiky," making direct comprehension challenging. Therefore, Implet aims to extract consecutive subsequences exhibiting high cumulative attribution scores. Additionally, to avoid trivial subsequences, we encourage longer subsequences, allowing minor gaps with near-zero attribution if necessary. More formally:

\begin{definition}
  Given a model $f$, a sample $\mathbf{x} = \{x_1, \dots, x_T\}$, and corresponding feature attributions $\mathbf{w} = \{w_1, \dots, w_T\}$, a subsequence
  \begin{equation}
    \mathbf{I}(l, r; \mathbf{x}, \mathbf{w}) \doteq \big(\{x_l, \dots, x_r\}; \{w_l, \dots, w_r\}\big)
    \label{eq:def}
  \end{equation}
  is considered an implet of $\mathbf{x}$ if its score $s$ satisfies:
  \begin{equation}
    s(l, r; \mathbf{x}, \mathbf{w}) \doteq \left(\sum_{i=l}^r |w_i|\right) + \lambda (r - l + 1) \ge \phi,
    \label{eq:def_cond_score}
  \end{equation}
  and its length is within predefined boundaries:
  \begin{equation}
     \ell_{\min} \le r - l + 1 \le \ell_{\max},
     \label{eq:def_cond_len}
  \end{equation}
  where $\lambda$ is a hyperparameter controlling subsequence length, and $\phi$ is a threshold.
\end{definition}

In Eq.~\ref{eq:def_cond_score}, the first term ensures high cumulative attribution within the subsequence, while the second term discourages excessively short segments. Additionally, the minimum length prevents single spikes in attribution being categorized as an Implet. In practice, we set $\lambda = 0.1$ and normalize the attribution scores $\mathbf{w}$. We also set $\phi = 1$, corresponding to a score threshold approximately one standard deviation above the mean of normalized attributions. Consistent with previous shapelet literature, we set length constraints as $\ell_{\min} = 3$ and $\ell_{\max} = \lfloor T/2 \rfloor$ \cite{shapelet}. 

To avoid overlapping subsequences, we sequentially extract the highest-scoring subsequence starting from each candidate location $l$. After identifying an implet, we update the search to resume from the subsequence’s endpoint. Additionally, candidate positions with attributions below $\phi$ are pruned to accelerate extraction. The detailed extraction algorithm is provided in Alg.~\ref{alg:implet_extraction}. The algorithm first find the first time step which have an attribution higher than $phi$, and then find a valid end location that maximize the score in Eq.~\ref{eq:def_cond_score}. This subsequence is added to the implet list. It then finds the next time step after the previous implet with an attribution higher than $\phi$, and repeat. Since the input attribution $\mathbf{w}^C$ is w.r.t class $C$, the result implets will also be correspond to $C$. The complexity of extracting implets for a single sample is $O(T)$.

\subsection{Coh-implet (Implet Clusters)}

To further enhance conciseness, we propose to cluster similar implets together, and report the centroids as succinct explanations. We name the centroids as \textbf{Coh-Implets}. Similar to Implets, Coh-Implets are 2-dimensional subsequences, of which the first dimension contains feature values, and the second dimension contains attributions. The difference is that a Coh-Implet is a representative centroid of a cluster of Implets.

 We achieve this via a modified $k$-means algorithm. The clustering distance metric is the two-dimensional dependent Dynamic Time Warping (DTW) distance \cite{dtw-ndim}, and cluster centroids are computed using DTW Barycenter Averaging (DBA) \cite{dba}. We choose DTW over sliding-window distance (used by shapelet algorithms) is to account for the variable subsequence length of implets. DTW also has the added benefit of being more suitable for post-hoc explanations, as network operations such as pooling allow models to perceive slightly stretched subsequences as equivalent patterns. Furthermore, we incorporate attribution information into the DTW computation, recognizing that identical subsequence values might influence the model differently depending on temporal context or correlated features. Thus, the two-dimensional (feature plus attribution) dependent DTW provides a more faithful measure for subsequence similarity in post-hoc explanations.

To automatically determine the optimal number of clusters $k$, we compute silhouette scores based on the two-dimensional dependent DTW metric. The full clustering algorithm is outlined in Alg.~\ref{alg:implet_cluster}, where $d_\mathrm{dtw}$ denotes the two-dimensional dependent DTW distance, and $\mathrm{Silhouette}_{\mathrm{dtw}}$ is the corresponding silhouette metric. Similar to $k$-means, the algorithm randomly initialize centroids by sample $k$ implets. It then assign each implet to the nearest cluster. The centroid are recomputed using DBA, and this process is repeated until convergence.

\section{Evaluations}

In this section, we demonstrate the faithfulness and interpretability of Implet through comprehensive evaluations. We begin by 
qualitatively analyze Implet explanations on widely used time-series datasets (Sect.~\ref{sec:eval-qual}). We then quantitatively assess the faithfulness of Implet by removing the identified subsequences and measuring their impact on model accuracy (Sect.~\ref{sec:eval-faithful}). Finally, we verify that after clustering, Coh-Implets also maintain faithfulness to the model (Sect.~\ref{sec:eval-cohort}).

Throughout this section, we evaluate Implet on two representative models: a simpler Fully Convolutional Network (FCN) \cite{fcn} and a more complex state-of-the-art model, InceptionTime \cite{inceptiontime}. We consider the most widely-used attribution methods, including DeepLIFT\footnote{\texttt{captum}'s DeepLIFT implement is incompatible iwth \texttt{tsai}'s InceptionTime due to its ReLU layers. Thus, we omit DeepLIFT explanations for InceptionTime.} \cite{deeplift}, Input$\times$Gradient \cite{inputxgrad}, LIME \cite{lime}, KernelSHAP \cite{shap}, Saliency \cite{saliency}, and Occlusion \cite{occlusion}. The models were implemented using the \texttt{tsai} package \cite{tsai}, attribution explainers with \texttt{captum} \cite{captum}, shapelets with \texttt{pyts} \cite{pyts}, and Dynamic Time Warping (DTW) / DBA computations with \texttt{dtaidistance} \cite{dtaidistance}.

\subsection{Qualitative Analysis}
\label{sec:eval-qual}

To illustrate the meaningfulness of Implet explanations, we tie Implets to domain knowledge and qualitatively analyze clusters of identified Implets on two popular time series datasets: GunPoint and Chinatown. These datasets contain clearly defined temporal patterns with intuitive explanations. We use FCN as the target model for both datasets, and select saliency as the attribution method due to its consistently strong performance in quantitative faithfulness evaluations (Sect.~\ref{sec:eval-faithful}).

\begin{figure}[t]
	\centering
	\begin{subfigure}[b]{0.24\textwidth}
		\centering
		\includegraphics[width=\linewidth]{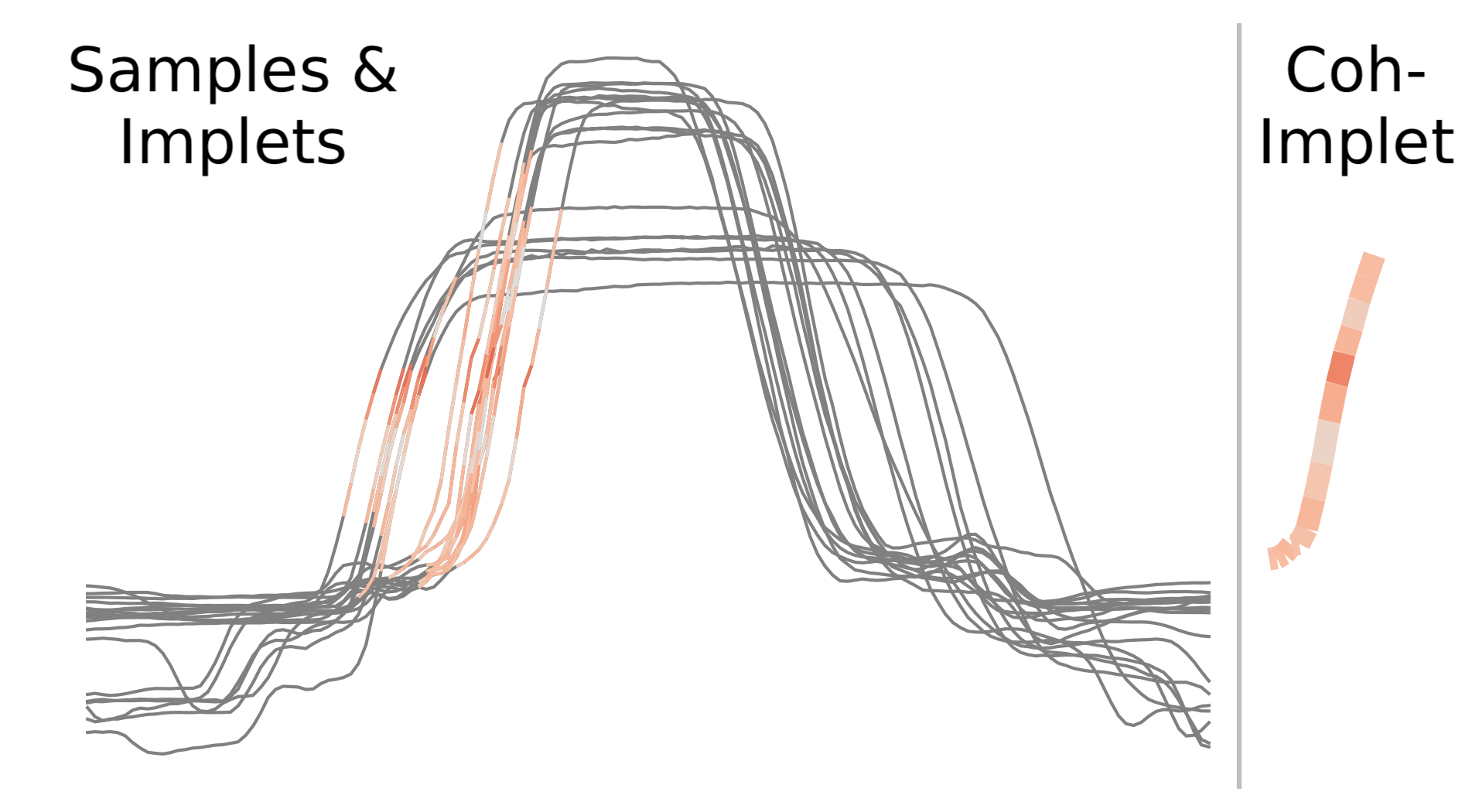}
		\caption{class \texttt{gun-draw}, cluster 1}
		\label{fig:gp_fig1a}
	\end{subfigure}
	\hfill
	\begin{subfigure}[b]{0.24\textwidth}
		\centering
		\includegraphics[width=\linewidth]{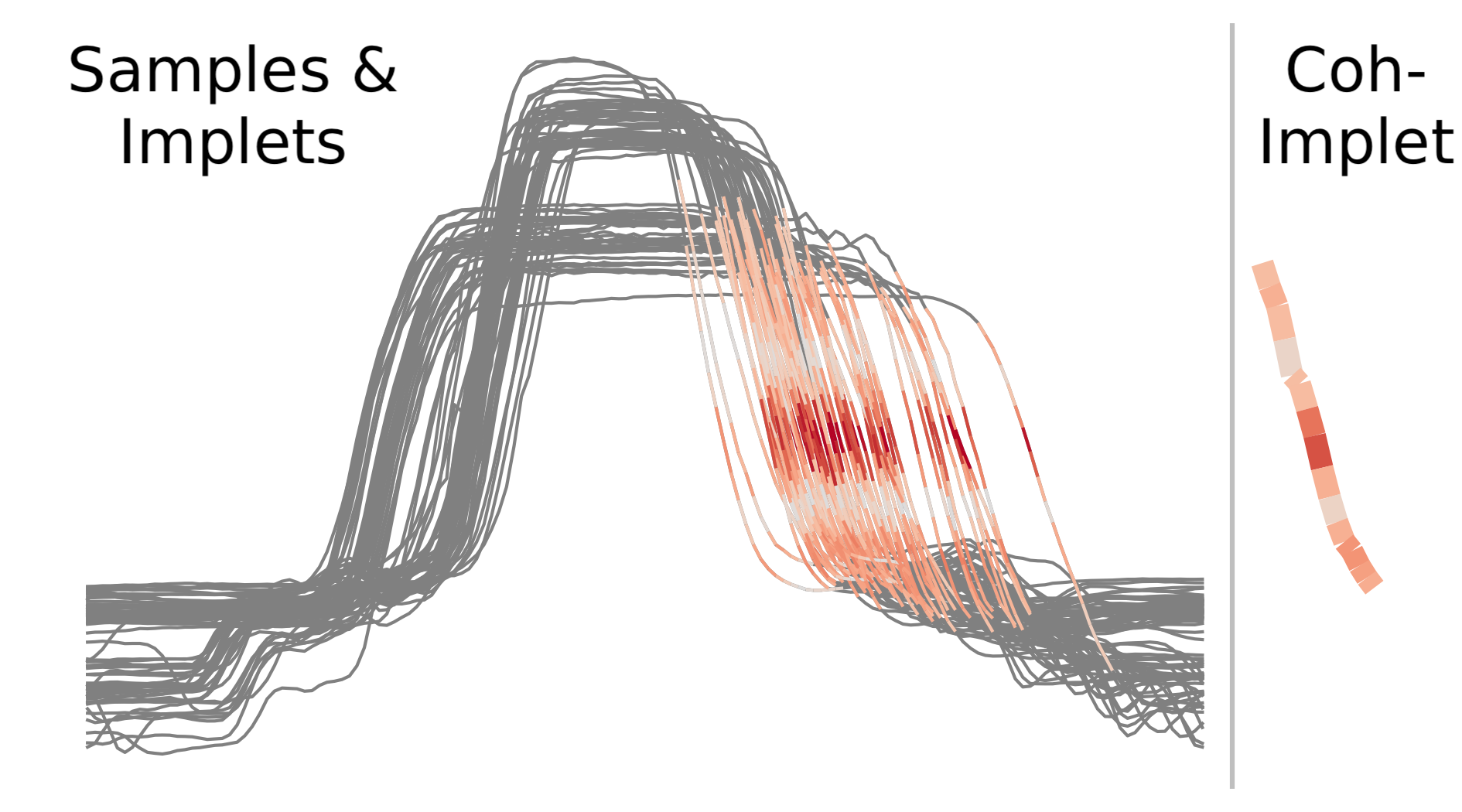}
		\caption{class \texttt{gun-draw}, cluster 2}
		\label{fig:gp_fig1b}
	\end{subfigure}
	\hfill
	\begin{subfigure}[b]{0.24\textwidth}
		\centering
		\includegraphics[width=\linewidth]{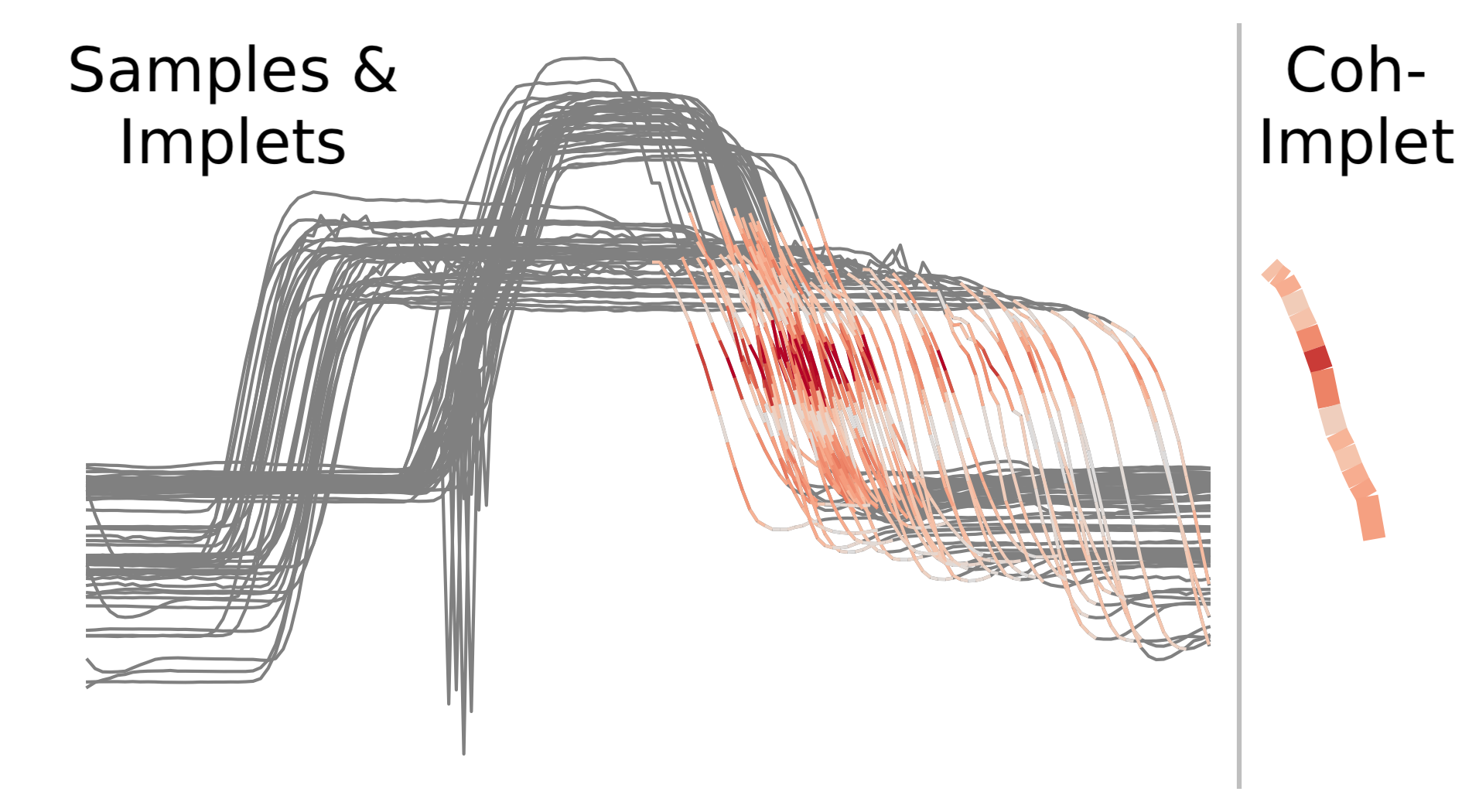}
		\caption{class \texttt{finger-pnt}, cluster 1}
		\label{fig:gp_fig1c}
	\end{subfigure}
	\hfill
	\begin{subfigure}[b]{0.24\textwidth}
		\centering
		\includegraphics[width=\linewidth]{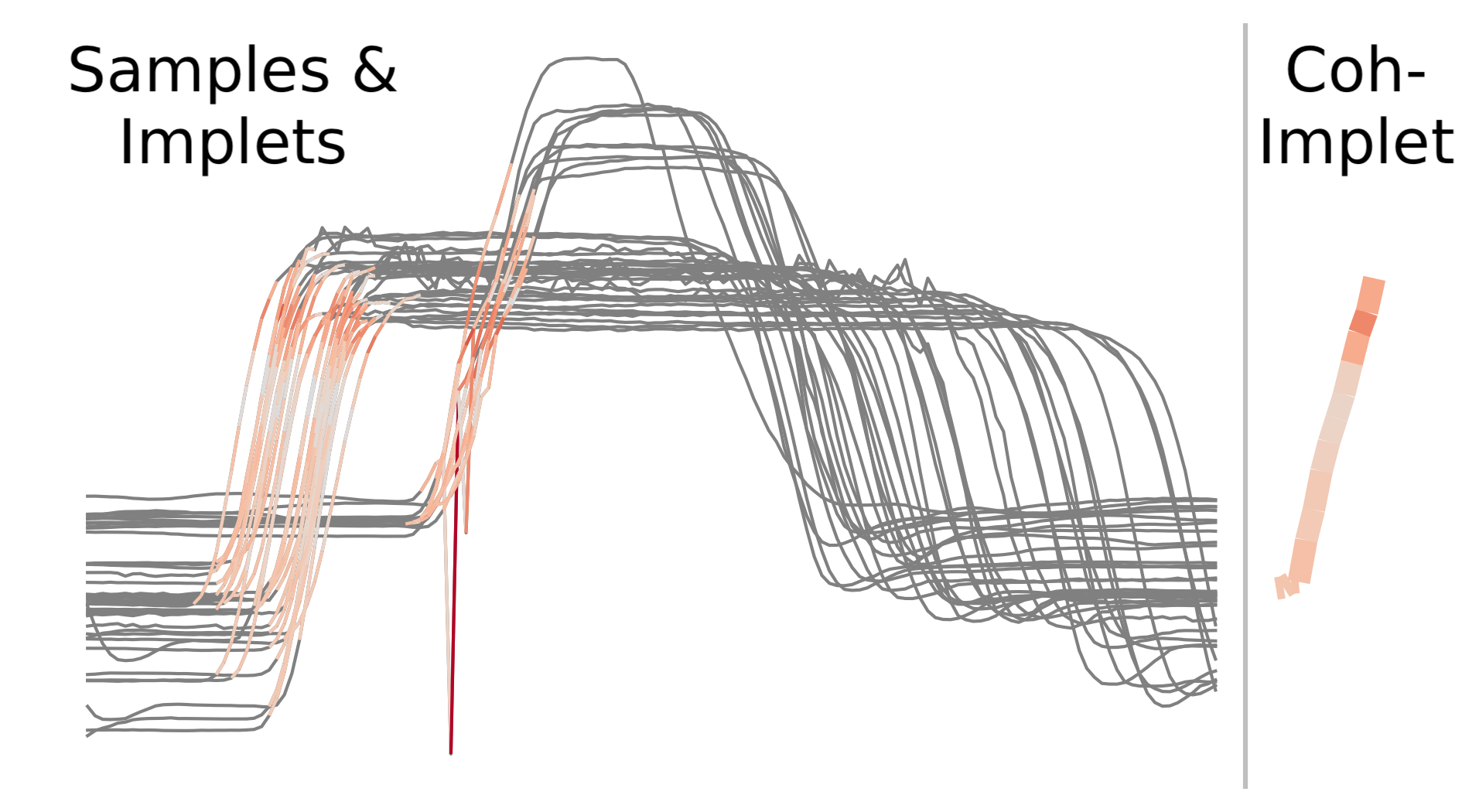}
		\caption{class \texttt{finger-pnt}, cluster 2}
		\label{fig:gp_fig1d}
	\end{subfigure}
	\caption{Implet cohort explanations for the GunPoint dataset. Each subfigure corresponds to one cohort, with highlighted regions representing the identified implets. The bold subsequence to the right of the dividing line denotes the Coh-Implet (cluster centroid) The color intensity represents attribution strength.}
	\label{fig:gp}
\end{figure}

\subsubsection{GunPoint}

The GunPoint dataset is a binary classification problem involving human activity recognition sensor data \cite{GP}. The two classes are \texttt{gun-draw} and \texttt{finger-pnt}. In the \texttt{gun-draw} class, participants begin with their hands positioned by their hips, draw a replica gun from a hip-mounted holster, point it toward a target for approximately one second, and then return the gun to the holster and their hands back to their sides. In the \texttt{finger-pnt} class, participants perform the same action using their index finger instead of a gun. The recorded data tracks participants' hand positions along the X-axis over time.

Fig.~\ref{fig:gp} visualizes the implet cohort explanations for the GunPoint dataset. Each class results in two distinct clusters: one highlighting upward (increasing) motions and the other capturing downward (decreasing) motions. The separation into these two clusters is consistent with physical intuition, as drawing motions involving a gun and a finger exhibit different acceleration and deceleration profiles. Also note that each cluster contains two types of visually distinct samples (e.g. in Fig.~\ref{fig:gp_fig1a}, there's ``taller" samples and ``wider" samples). The reason behind this is that the GunPoint dataset collects samples using two subjects (one male and one female) for both classes. Despite the overall shape differences of the two subjects, Coh-Implet recognizes the similarity of subsequences from the two subjects. Therefore, each cluster contains subsequences from \textit{both} subjects, presenting the most meaningful temporal subsequences influencing the model's classification decisions.

\begin{figure}[t]
	\centering
	\begin{subfigure}[b]{0.24\textwidth}
		\centering
		\includegraphics[width=\linewidth]{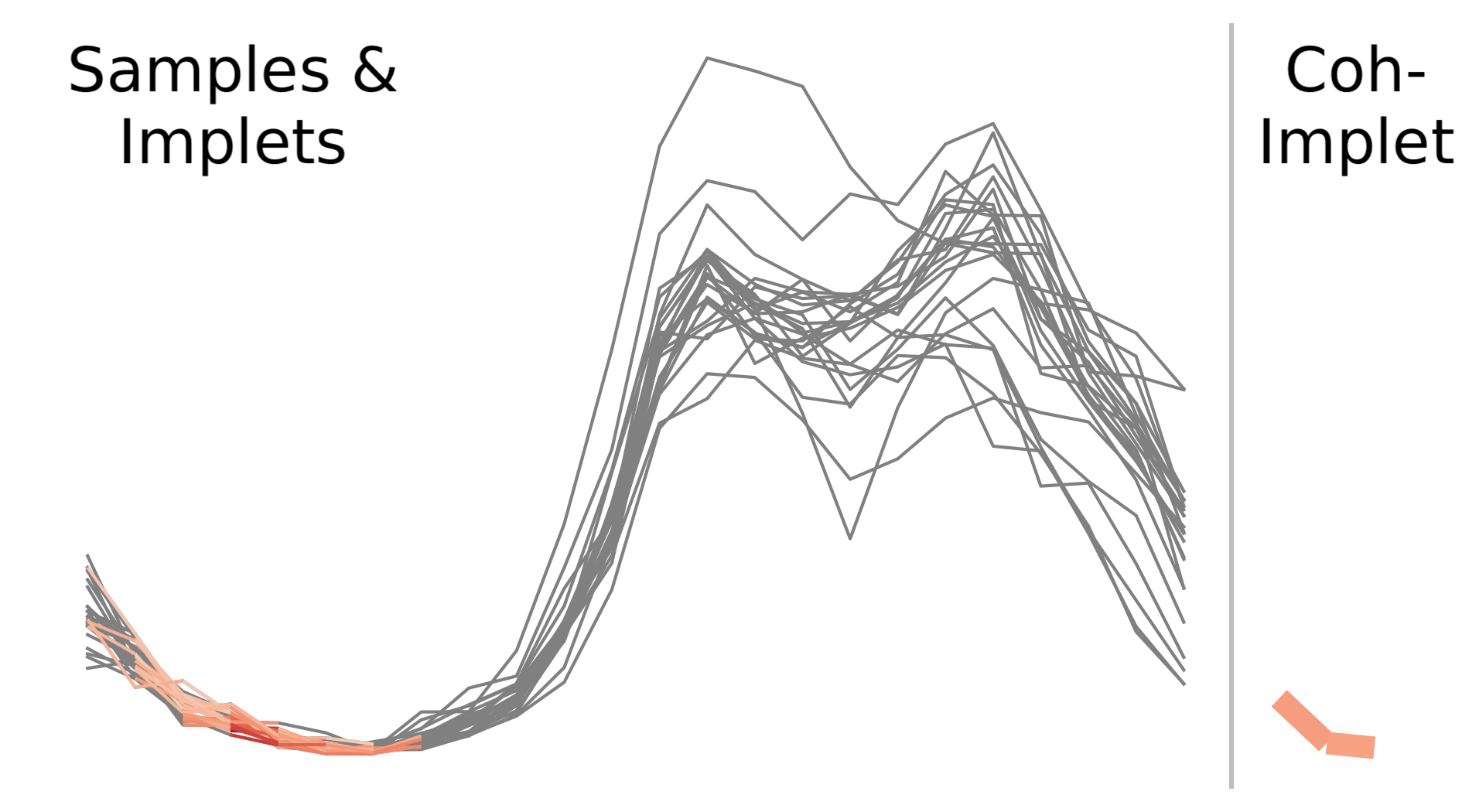}
		\caption{\texttt{weekend} cluster}
		\label{fig:ct_fig1a}
	\end{subfigure}
	\hfill
	\begin{subfigure}[b]{0.24\textwidth}
		\centering
		\includegraphics[width=\linewidth]{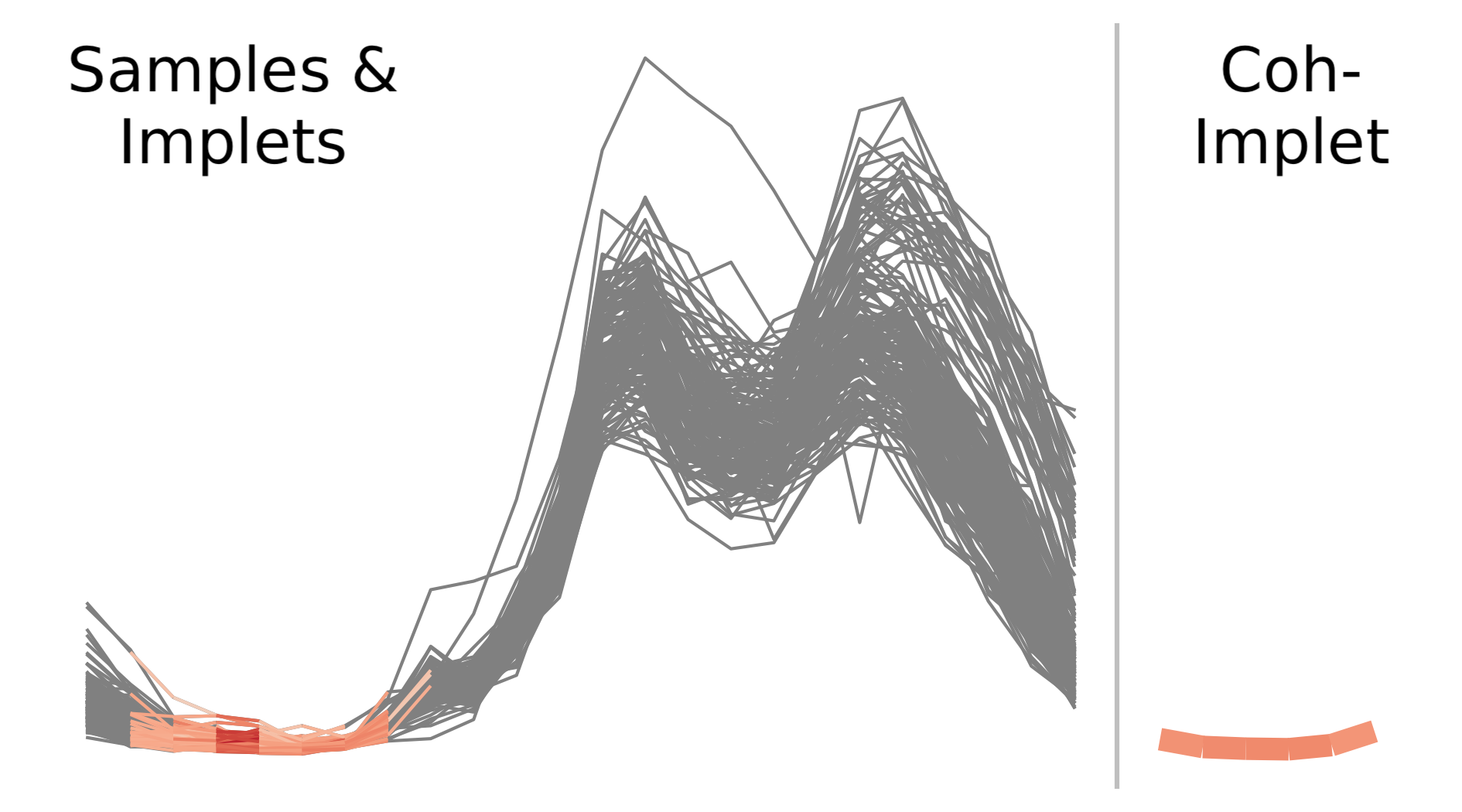}
		\caption{\texttt{weekday} cluster}
		\label{fig:ct_fig1b}
	\end{subfigure}
	\caption{Implet cohort explanations for the Chinatown dataset. Each figure represents one cohort, with highlighted regions denoting the identified implets. The bold subsequence to the right of the dividing line denotes the Coh-Implet (cluster centroid). The color intensity represents attribution strength.}
	\label{fig:ct}
\end{figure}

\subsubsection{Chinatown}

The Chinatown dataset comprises hourly pedestrian traffic data collected in Melbourne, Australia's Chinatown throughout 2017. Each sample spans 24 hours, with each time step recording the total pedestrian count in the corresponding hour. The classification task is to distinguish between weekend and weekday pedestrian traffic patterns.

Implets and Coh-Implets on the Chinatown dataset are visualized in Fig.~\ref{fig:ct}. Each class yields one main cluster, both focusing prominently on the early morning hours (1 AM to 6 AM). This aligns with intuition: weekend patterns exhibit higher pedestrian traffic during late-night hours compared to weekdays, where traffic tends to decline earlier. These findings validate that Implet successfully extracts human-interpretable temporal patterns consistent with our expectations.

\begin{figure}[t]
  \centering
  \includegraphics[width=0.6\linewidth]{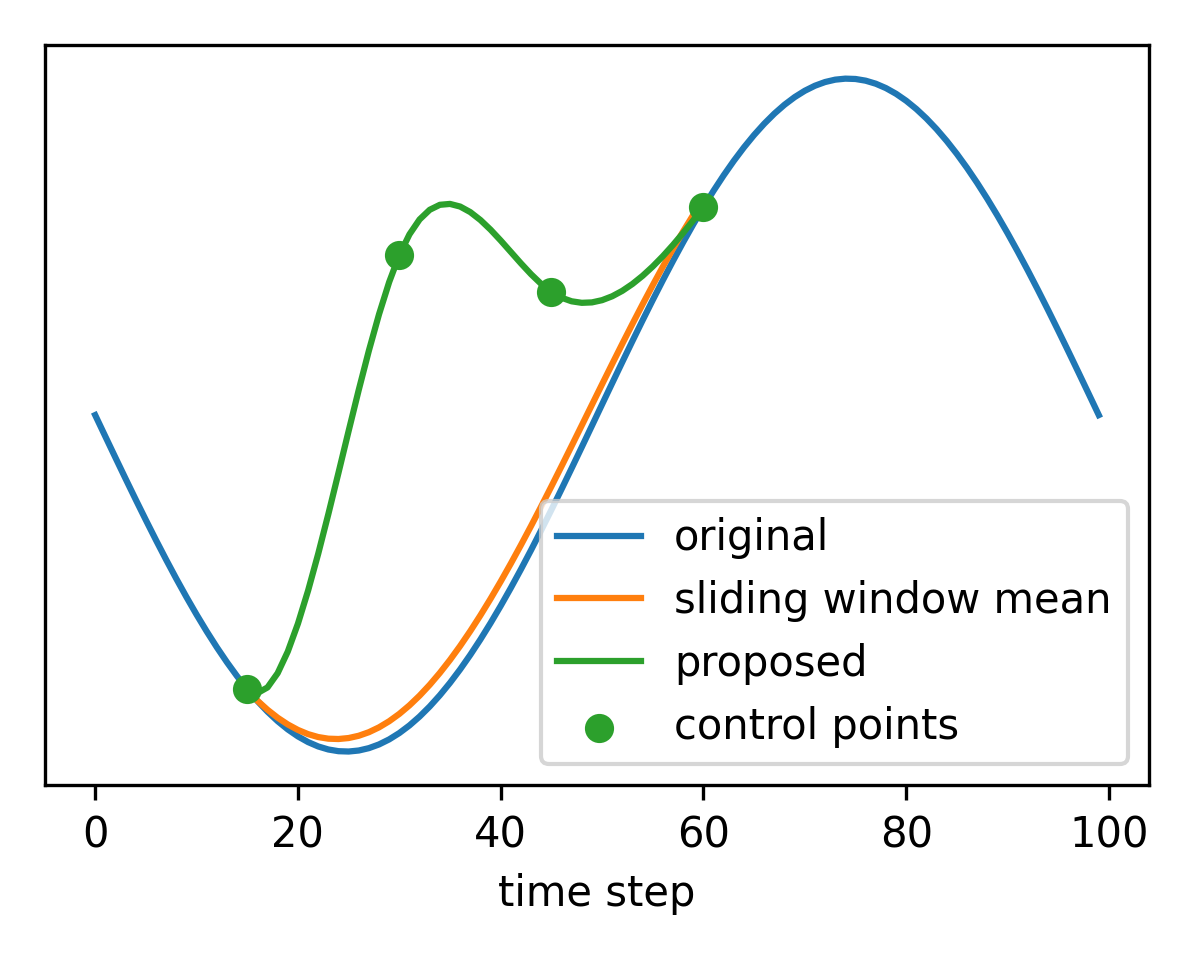}
  \caption{Example of the proposed subsequence removal scheme.}
  \label{fig:blur_demo}
\end{figure}

\begin{figure*}[t]
	\centering
	\begin{subfigure}{\textwidth}
		\centering
		\includegraphics[width=0.95\linewidth]{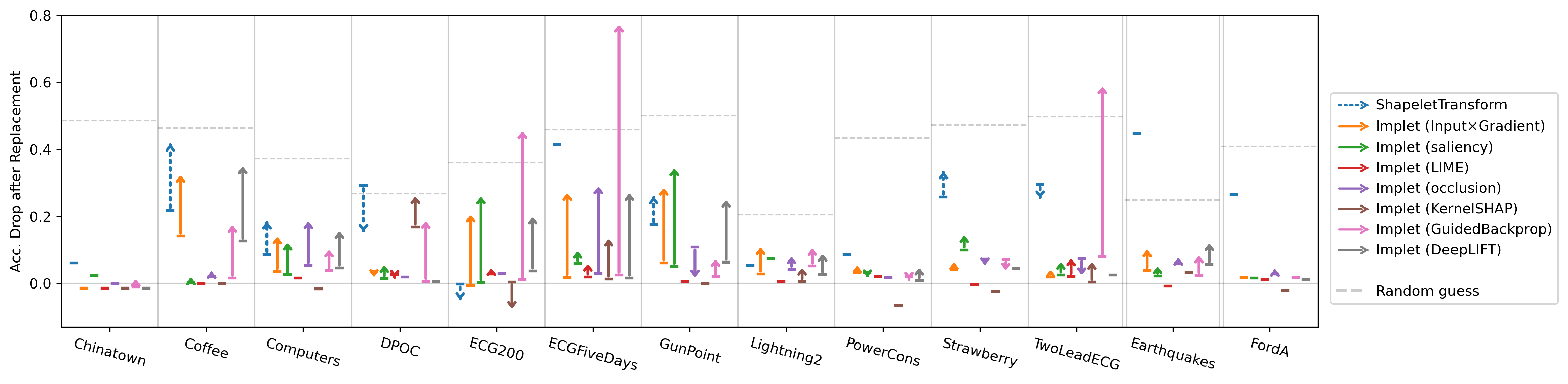}
		\caption{FCN}
		\label{fig:blur_fcn_arrow}
	\end{subfigure}
	\begin{subfigure}{\textwidth}
		\centering
		\includegraphics[width=0.95\linewidth]{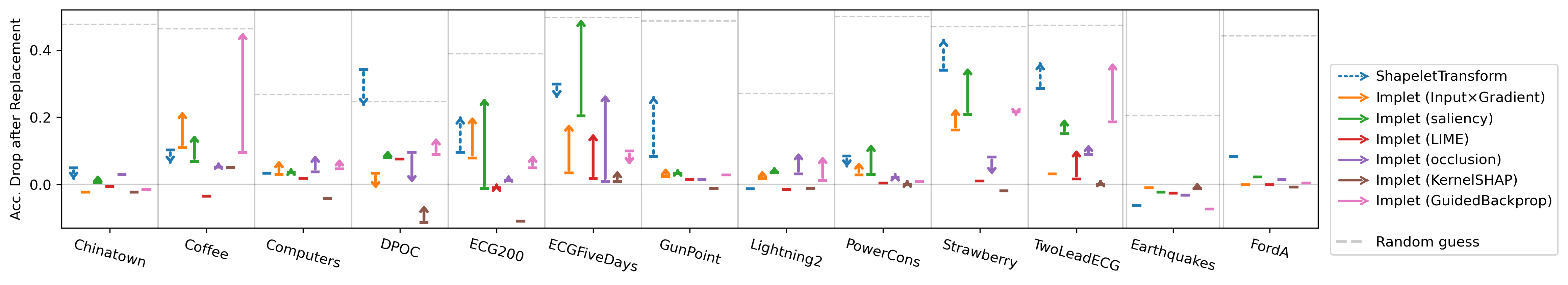}
		\caption{InceptionTime}
		\label{fig:blur_it_arrow}
	\end{subfigure}
	\caption{Faithfulness evaluation of implets with different attribution methods, compared against the baseline ShapeletTransform. Arrows indicate the accuracy drop from removing identified subsequences (arrow tip) versus removing random subsequences of equal length (arrow tail). e.g. an arrow pointing from 0.2 to 0.4 represents removing random subsequence causes an accuracy drop of 0.2, while removing the explainer output causes an accuracy drop of 0.4. Longer, upward arrows indicate more faithful explanations. Short horizontal lines indicate negligible differences between random and identified subsequence removal. Dashed arrow represent ShapeletTransform for clarity. Horizontal dashed lines correspond to the accuracy drops that are equivalent to random guess. The last two dataset differ from the rest as \texttt{Earthquaks} is event-based and \texttt{FordA} is frequency-based.}
	\label{fig:blur_arrow}
\end{figure*}


\begin{figure*}[t]
	\centering
	\begin{subfigure}{\textwidth}
		\centering
		\includegraphics[width=0.95\linewidth]{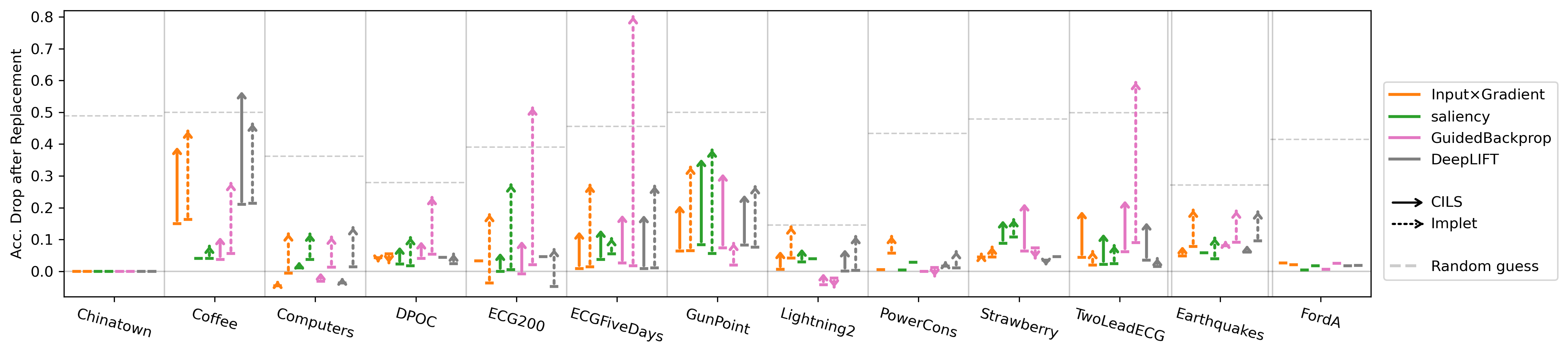}
		\caption{FCN}
		\label{fig:cohort_fcn}
	\end{subfigure}
	\begin{subfigure}{\textwidth}
		\centering
		\includegraphics[width=0.95\linewidth]{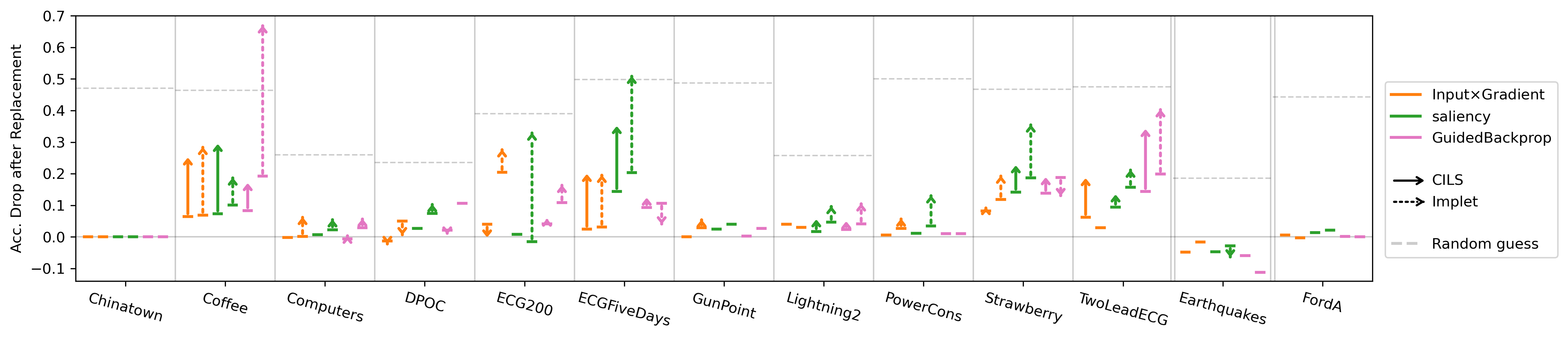}
		\caption{InceptionTime}
		\label{fig:cohort_it}
	\end{subfigure}
	\caption{Faithfulness evaluation comparing Implets (dashed arrows) and Implet-Centroid-Like Subsequences (ICLS, solid arrows). Arrows indicate the accuracy drop from removing identified subsequences (arrow tip) versus removing random subsequences of equal length (arrow tail). Longer, upward arrows indicate more faithful explanations. Short horizontal lines indicate negligible differences between random and identified subsequence removal. Horizontal dashed lines correspond to the accuracy drops that are equivalent to random guess. The last two dataset differ from the rest as \texttt{Earthquaks} is event-based and \texttt{FordA} is frequency-based.}
	\label{fig:cohort}
\end{figure*}

\subsection{Implet Faithfulness}
\label{sec:eval-faithful}

To quantitatively evaluate the faithfulness of Implet explanations, we conduct an ablation analysis. The core idea is to identify subsequences highlighted by Implet and subsequently remove them from the input samples. If the identified subsequences (Implets) genuinely reflect the model’s decision-making, then their removal should cause a significantly greater drop in accuracy compared to randomly removing subsequences of equivalent length. Thus, we define faithfulness as the mean accuracy drop after ``removing" the subsequences from all samples.

However, effectively removing subsequences from time series data poses challenges. Many time series models are sensitive to sudden changes or abrupt transitions \cite{abrupt1, abrupt2}. Simple removal methods such as zero-filling may create artificial discontinuities, inadvertently affecting model behavior. Other naive methods like sliding-window averages or Gaussian removal are also inadequate, as they have limited effects on smooth yet crucial subsequences (see Fig.~\ref{fig:blur_demo}). GunPoint implets in Fig.~\ref{fig:gp} are examples of smooth subsequences that are difficult for sliding-window averages to remove. Other baselines include filling with sample mean \cite{baseline-mean}, reverse the time order of the subsequence \cite{baseline-reverse},  and learning the most useful replacement from a different dataset \cite{baseline-learn}. All of these baselines are subject to abrupt transitions. Thus, to address these issues, we propose a simple, specialized removal procedure via randomized polynomials:

\begin{enumerate}
  \item Determine the number of control points with $max(\lceil L/10\rceil, 2)$, where $L$ is the length of the subsequence;
  \item Assign each control point a random i.i.d. value drawn from a distribution with the same mean and standard deviation as the original sample.
  \item Add the subsequence starting and ending values to the control points. Then interpolate a polynomial connecting the control points. Also match the polynomial's gradients at the start and end of the subsequence with the gradients on the original sample.
\end{enumerate}

Fig.~\ref{fig:blur_demo} shows an example of this procedure. It ensures smooth replacement signals, minimizing unintended model reactions.

We perform our evaluations using 13 commonly used binary classification datasets from the UCR Time Series Archive \cite{ucr} \footnote{In Fig.~\ref{fig:blur_arrow} and \ref{fig:cohort}, \texttt{DPOC} denotes \texttt{DistalPhalanxOutlineCorrect}.}. Model performances can be found in Appendix~\ref{app:acc}. Both FCN and InceptionTime achieve high accuracy on all tasks, though on certain datasets InceptionTime exhibits slight overfitting due to a large parameter size and the limited dataset size. We evaluate across multiple attribution methods previously discussed. Additionally, we benchmark our approach against ShapeletTransform \cite{shapelet-transform} as a \textit{model-agnostic} baseline.\footnote{We initially attempted to compare against LASTS \cite{lasts}. However, available implementations failed to identify meaningful subsequences in most datasets, as perturbations in its latent space rarely flipped predictions, resulting in nonexistent explanations. Thus, LASTS is excluded from our comparison.}

Fig.~\ref{fig:blur_arrow} summarizes the results from our subsequence removal experiments. Removing subsequences identified by ShapeletTransform often leads to notable accuracy drops, indicating that the shapelets generally contain important model-focused features. However, because shapelets tend to be lengthy and noisy, removing random subsequences of equal length typically results in a comparable accuracy drop. On several occasions, random removal leads to an even greater accuracy reduction, suggesting that shapelets sometimes include less informative segments. This matches our expectation, as though Shaplets learns discriminative subsequences, they may not match what the target model has learned.

In contrast, Implet subsequences demonstrate consistent superiority: removing implets almost always leads to significantly larger accuracy drops compared to random subsequence removal. Among attribution methods, Implet performs particularly well when paired with Saliency, Input$\times$Gradient and DeepLIFT, producing consistently high-quality explanations. GuidedBackprop perform well with simpler models such as FCN but are less effective with deeper architectures like InceptionTime. This might due to GuidedBackprop zero-ing out negative gradients, causing it to become oblivious to higher layer weights \cite{saliency-sanity}. The two perturbation-based approach, LIME and KernelSHAP, generally yield sparse, fragmented attributions caused by their built-in sparsity constraint or regularization term \cite{lime, shap}. Sparsity might be desirable for a standalone explainer, but it hinder effective subsequence identification, resulting in poorer Implet performance. 

Also, we observe that all methods are less effective on InceptionTime than FCN. We hypothesize that the reason is InceptionTime being more powerful than FCN. Since we only remove one segment in each perturbed sample (equivalent to removing one feature in a tabular task), not all discriminative information is removed. With a much larger receptive field, InceptionTime could overcome perturbation by inferring the label from unmodified segments. This is more significant on simpler tasks such as \texttt{GunPoint}. However, the difference between removing Implets and removing random segments is still non-trivial in most tasks, demonstrating that Implet does find the important segments used by the model. 

We note, however, that certain datasets—particularly \texttt{Earthquakes} and \texttt{FordA}—pose difficulties for subsequence-based explainers, including both Implet and ShapeletTransform. These datasets are either frequency-oriented or event-based, which limits the effectiveness of subsequence explanations \cite{exp_space}. Additionally, the \texttt{Chinatown} dataset presents challenges due to its short time series length (24 steps), resulting in very short identified implets (typically length 3), which makes our smooth removal method less impactful. We consider this lack of accuracy drop is more related to the smooth removal than the effectiveness of Implets. We analyze with different removals method on these tasks and obtain better results, and we report them in Appendix~\ref{app:alt-removal}.

\subsection{Cohort Explanation: Coh-Implet}
\label{sec:eval-cohort}

In the previous sections, we demonstrated that implets provide faithful and interpretable subsequence explanations. Here, we investigate whether the cohort explanation framework—particularly the centroids generated by clustering implets—is also faithful to the underlying model. To evaluate this, we design an experiment using the same datasets, models, and attribution methods as previously in Sect.~\ref{sec:eval-faithful}. The evaluation pipeline is structured as follows:

\begin{enumerate}
    \item Split each dataset evenly into two subsets: the \textit{Implet extraction} subset and the \textit{Implet evaluation} subset.
    \item Extract implets from the \textit{Implet extraction} subset using Algorithm~\ref{alg:implet_extraction}, and then cluster them using Algorithm~\ref{alg:implet_cluster} to obtain the Coh-Implets. 
    \item In the \textit{Implet evaluation} subset, find subsequences that are similar to the identified Coh-Implets. For each Coh-Implet, identify subsequences that best match the centroid shape using 1-D DTW solely based on the feature values.  We refer to these subsequences as \textit{Coh-Implet-Like Subsequences} (CILS).
    \item Evaluate the impact of removing the identified CILS on model accuracy, following the ablation approach described in Sect.~\ref{sec:eval-faithful}.
    \item As a baseline, extract implets directly from the \textit{Implet evaluation} subset and measure accuracy drops using the same removal approach.
\end{enumerate}

The results of this evaluation are presented in Fig.~\ref{fig:cohort}, comparing the accuracy drops caused by removing CILS versus removing the original implets. Due to space constraints, we show results only for the attribution methods exhibiting consistently strong performance: Input$\times$Gradient, Saliency, GuidedBackprop, and DeepLIFT. Across most datasets, we observe that removing ICLS results in accuracy reductions comparable to removing implets, \textit{even though no attribution information is used in obtaining CILS}. This indicates that cohort centroids capture model-relevant information and thus faithfully represent the underlying model behavior.

Note that in identifying CILS, we deliberately omit the attribution dimension and consider only the raw time series values. We want to prevent bias towards higher attribution areas, since the implet centroids' attribution dimension is naturally high. Instead, we want to verify that the shape of the implet centroids alone remain meaningful and faithful to model behavior. Therefore, this is a stronger results than if we find CILS using both dimensions. If we were to use both dimensions, their faithfulness would be between 1D CILS and real implets. Such results are reported in Appendix~\ref{app:cohort-2d}.

Additionally, removing CILS is slightly less effective than removing implets, underscoring the importance of including attribution information during Implet extraction and clustering. Without attribution, clusters may combine subsequences that appear similar in value but differ significantly in terms of model impact, reducing overall explanation effectiveness.

\section{Conclusion}

In this work, we introduced \textit{Implet}, a novel post-hoc subsequence explainer designed to enhance interpretability of time series models by identifying critical temporal segments derived from feature attributions. Unlike traditional shapelet-based approaches, Implet is explicitly model-aware, generating concise and faithful explanations that directly reflect the model's decision-making process. Additionally, we proposed a cohort explanation framework to cluster similar implets, producing higher-level, interpretable summaries. Through extensive qualitative and quantitative evaluations, we demonstrated that Implet significantly improves the faithfulness, conciseness, and clarity of explanations across a variety of datasets. Our approach effectively bridges the gap between fine-grained feature attribution methods and human-interpretable temporal explanations, paving the way toward more transparent and trustworthy time series AI systems.

In our experiments, we focused primarily on binary classification tasks, as they are more straightforward for interpretation and visualization purposes. However, Implet can be naturally extended to higher-dimensional data by generalizing each implet from 2-dimensional (value and attribution) to $2n$-dimensional, where $n$ represents the dimensionality of the original input sample. A promising direction for future research involves automatically identifying and removing dimensions that contribute minimal explanatory information, further improving the conciseness and usability of implets in complex, high-dimensional scenarios.
 
\appendix
\subsection{Model Performance}
\label{app:acc}

Table~\ref{tab:acc} shows the performance of the two models on the UCR datasets. InceptionTime is more expressive than FCN, though due to limited dataset size, it exhibits overfitting in some datasets and yield a slightly worse performance than FCN. It can be improved by hyperparameter tuning, but this is beyond the scope of this paper as we are focusing on explanation.

\begin{table}
  \centering
  \small
  \begin{tabularx}{0.8\linewidth}{c *{2}{Y}}
  \toprule
  & FCN & InceptionTime \\
  \midrule
  \texttt{Chinatown} & 98.54\% & 97.67\% \\
  \texttt{Coffee} & 96.43\% & 96.43\% \\
  \texttt{Computers} & 87.20\% & 76.80\% \\
  \texttt{DPOC} & 76.81\% & 74.64\% \\
  \texttt{ECG200} & 86.00\% & 89.00\% \\
  \texttt{ECGFiveDays} & 95.94\% & 99.77\% \\
  \texttt{GunPoint} & 100.00\% & 98.67\% \\
  \texttt{Lightning2} & 70.49\% & 77.05\% \\
  \texttt{PowerCons} & 93.33\% & 99.77\% \\
  \texttt{Strawberry} & 97.30\% & 100.00\% \\
  \texttt{TwoLeadECG} & 99.74\% & 97.37\% \\
  \texttt{Earthquakes} & 74.82\% & 70.50\% \\
  \texttt{FordA} & 90.91\% & 94.24\% \\
  \bottomrule
  \end{tabularx}
  \caption{Prediction accuracy of the two models on the UCR datasets.}
  \label{tab:acc}
\end{table}

\begin{figure*}[t]
	\centering
	\begin{subfigure}{\textwidth}
		\centering
		\includegraphics[width=0.95\linewidth]{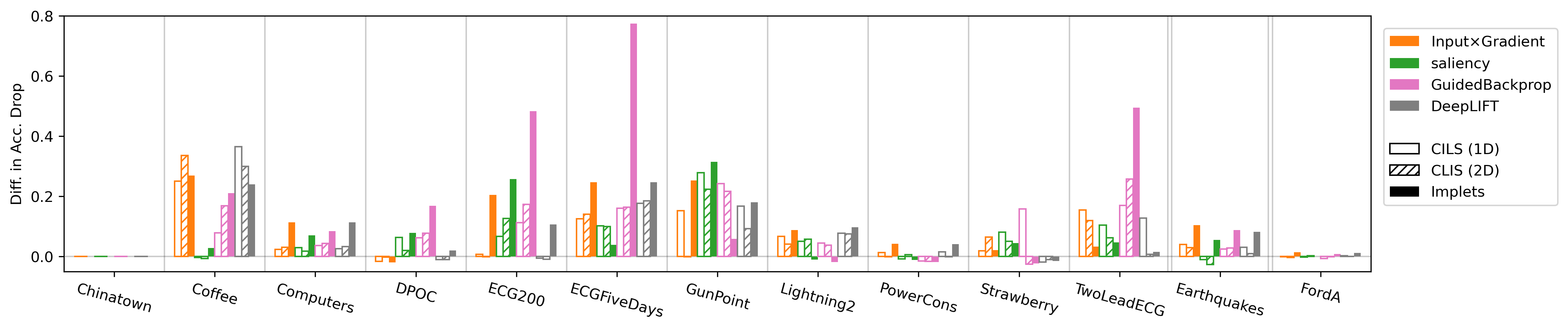}
		\caption{FCN}
		\label{fig:cohort_2d_fcn}
	\end{subfigure}
	\begin{subfigure}{\textwidth}
		\centering
		\includegraphics[width=0.95\linewidth]{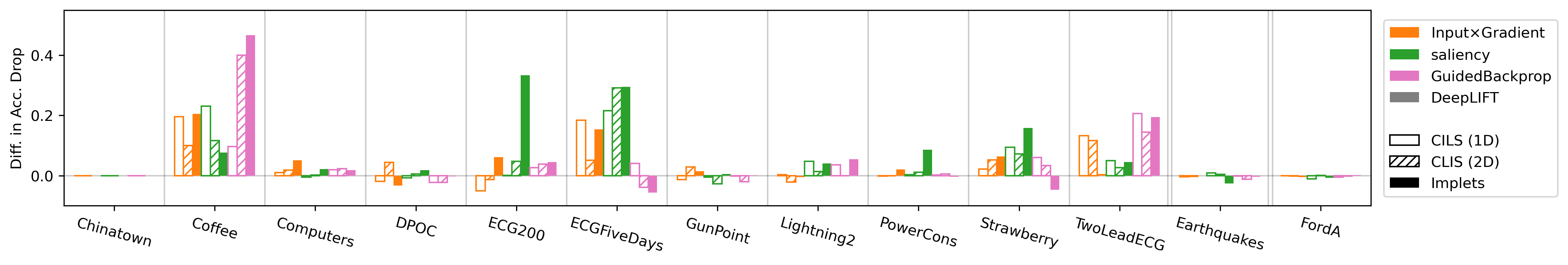}
		\caption{InceptionTime}
		\label{fig:cohort_2d_it}
	\end{subfigure}
	\caption{Faithfulness evaluation comparing (1) Implets, (2) CILS using feature dimension only, and (3) CILS using both feature and attribution dimensions. The height of the bars represent the difference in accuracy drop between removing identified subsequences versus removing random subsequences of equal length. Values are equivalent to arrow lengths in Fig.~\ref{fig:cohort}. Higher values indicate more faithful explanations. The last two dataset differ from the rest as \texttt{Earthquaks} is event-based and \texttt{FordA} is frequency-based.}
	\label{fig:cohort_2d}
\end{figure*}

\subsection{Implet Faithfulness with Alternative Removal}
\label{app:alt-removal}

In Sect.~\ref{sec:eval-faithful}, we observe that neither ShapeletTransform nor Implet achieve high faithfulness score on three datasets: \texttt{Chinatown}, \texttt{Earthquakes} and \texttt{FordA}. We hypothesize each has a different reason. \texttt{Chinatown} samples are short (24 steps), causing the Implets to also be short, limiting the effectiveness of the proposed smooth subseuqnces removal method. For the later two datasets, \texttt{Earthquakes} is an event-based task, and multiple events are needed to correctly classify \cite{ucr}. \texttt{FordA}'s primary features are frequency-based and appears multiple times in each sample. On these two datasets, removing one subsequence has limited effects. To verify our hypothesis that Implets are faithful to the target models on these datasets, and the lacking results in Fig.~\ref{fig:blur_arrow} are due to experiment designs, we conduct the following additional evaluations:
\begin{itemize}
  \item On \texttt{Chinatown}, instead of using the smooth subsequence removal procedure, we fill the target subsequence with the sample mean;
  \item On \texttt{Earthquakes} and \texttt{FordA}, instead of removing each subsequence individually, we find all Implets from each sample, remove all of them and then test for accuracy. The random removal baseline is also changed to remove the same number of random subsequences with equivalent lengths.
\end{itemize}
The results are reported in Fig.~\ref{fig:blur_alt}. We observe better faithfulness on \texttt{Chinatown} and \texttt{Earthquakes}, matching our expectations. Note that ShapeletTransform's results do not change significantly compared to the test in Sect.~\ref{sec:eval-faithful}. Additionally, on \texttt{Chinatown}, saliency and GuidedBackprop implets causes the model performance to degrade to be much lower than random guess. This suggests that Implets found with these two explainers are crucial to model predictions, and perturbing them would drastically affect model behavior. 

On \texttt{FordA}, Input$\times$Gradient, occlusion and GuidedBackprop achieves decent performance on InceptionTime. All subsequences explainers are still less effective on FCN. We conclude that the frequency features are challenging for subsequence explainers.

\begin{figure}[t]
  \centering
  \includegraphics[width=\linewidth]{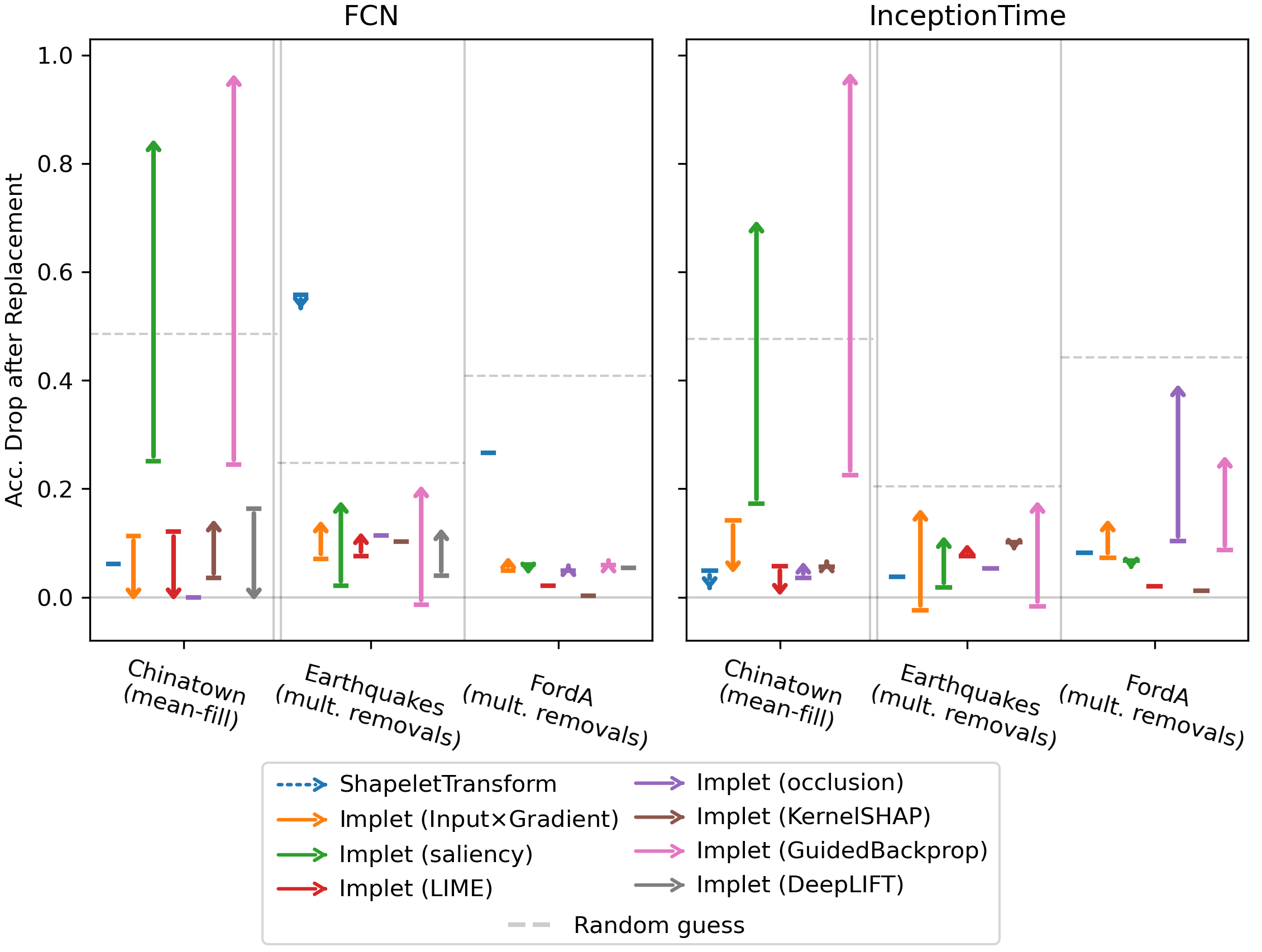}
  \caption{Alternative faithfulness evaluation on three datasets that do not perform well in Fig.~\ref{fig:blur_arrow}. \texttt{Chinatown} uses mean-fill as the subsequence removal method, and \texttt{Earthquakes} and \texttt{FordA} removes multiple subsequences from each sample. Arrows indicate the accuracy drop from removing identified subsequences (arrow tip) versus removing random subsequences of equal length (arrow tail). Longer, upward arrows indicate more faithful explanations. }
  \label{fig:blur_alt}
\end{figure}

\subsection{Evaluating Coh-Implets using Both Dimensions}
\label{app:cohort-2d}

In Sect.~\ref{sec:eval-cohort}, we evaluate the quality of Coh-Implets by finding Coh-Implet-Like subsequence (CILS). We achieve this by looking for subsequences that are similar to Coh-Implets in the feature value dimension. In this section, we perform the same evaluate, albeit we use both the feature and the attribution dimensions in finding CILS. 

Fig.~\ref{fig:cohort_2d} shows the 2D results. Intuitively, using both dimensions when finding CILS should yield a higher change in accuracy drop, since Coh-Implets naturally have higher attribution values. This will cause CILS to be biased towards higher attribution areas, causing 2D CILS to be more similar to Implets. We observe that in majority of the datasets and explainer combinations, the difference in accuracy drop follows 1D CILS $<$ 2D CILS $<$ Implets, which matches our expectation. This further motivates the usage of the attribution dimension in Implets. 

However, the performances of 1D CILS are not far from 2D CILS or Implets, barring a few exceptions with GuidedBackprop. This shows that after clustering Coh-Implets correctly captures the most relevant information to the target model.

\bibliographystyle{IEEEtran}
\bibliography{implet}

\end{document}